\newif\iftr 
\trtrue 
 
\documentclass[conference, 10pt]{IEEEtran}
\IEEEoverridecommandlockouts
\usepackage{cite}
\usepackage{amsmath,amssymb,amsfonts}
\usepackage{graphicx}
\usepackage{textcomp}
\usepackage{xcolor}
\usepackage{bm}
\usepackage{cases}
\usepackage[skip=1pt]{caption}
\usepackage[ruled,vlined]{algorithm2e} 
\usepackage[makeroom]{cancel} 

\makeatletter
\def\BibTeX{{\rm B\kern-.05em{\sc i\kern-.025em b}\kern-.08em
    T\kern-.1667em\lower.7ex\hbox{E}\kern-.125emX}}
    
\newcommand{\Rbb}{\mathbb{R}}
\mathchardef\myhyphen="2D

\newcommand{\xhat}{\hat{\xvec}}
\newcommand{\xhatinit}{\hat{\xvec}^0}
\newcommand{\xhatt}{\hat{\xvec}^t}

\newcommand{\xhattp}{\hat{\xvec}^{t+1}}
\newcommand{\dx}{{\dxvec}}
\newcommand{\dxt}{{\dxvec}^t}
\newcommand{\xtilde}{\Tilde{\xvec}}
\newcommand{\xtildet}{\Tilde{\xvec}^t}
\newcommand{\xtildetp}{\Tilde{\xvec}^{t+1}}

\newcommand{\cocoa}{\textsc{CoCoA}}
\newcommand{\cola}{\textsc{CoLa}}

\newcommand{\matr}[1]{\bm{#1}}
\newcommand{\Amat}{\matr{A}}

\newcommand{\Cmat}{\matr{C}}
\newcommand{\Dmat}{\matr{D}}

\newcommand{\Mmat}{\matr{M}}

\newcommand{\Qmat}{\matr{Q}}

\newcommand{\Umat}{\matr{U}}
\newcommand{\Vmat}{\matr{V}}
\newcommand{\Wmat}{\matr{W}}
\newcommand{\Gmat}{\matr{G}}
\newcommand{\Lambmat}{\matr{\Lambda}}
\newcommand{\avec}{\matr{a}}
\newcommand{\cvec}{\matr{c}}
\newcommand{\dxvec}{\Delta\matr{x}}
\newcommand{\vvec}{\matr{v}}
\newcommand{\vbar}{\bar{\matr{v}}}

\newcommand{\xvec}{\matr{x}}
\newcommand{\yvec}{\matr{y}}
\newcommand{\zvec}{\matr{z}}

\newcommand{\Imat}{\matr{I}}

\newcommand{\rank}{\text{Rank}}

\newcommand{\eye}[1]{\Imat_{#1}}

\newcommand{\zerobf }{\boldsymbol 0}

\newcommand{\Ebb}{\mathbb{E}}

\DeclareMathAlphabet{\mymathbb}{U}{BOONDOX-ds}{m}{n}

\newcommand{\T}{^\mathrm{T}}

\newcommand{\p}{^+}

\newcommand{\norm}[1]{\big\lVert#1\big\rVert_2}

\let\oldin\in
\renewcommand{\in}{{\,\oldin\,}}

\let\oldnotin\notin
\renewcommand{\notin}{{\,\oldnotin\,}}

\newcommand{\tr}{\operatorname{tr}}

\newcommand{\irange}{{i=1,\,\dots,\,n}}

\newcommand{\Gauss}[1]{\mathcal{N}(0,\,\eye{#1})}

\newcommand{\Aplusmat}{\begin{bmatrix}\Amat_1\p\\\vdots\\\Amat_K^+\end{bmatrix}}
\newcommand{\Aplusmathorz}{[\Amat_1\p;\cdots;\Amat_K^+]}

\long\def\/*#1*/{}

\newcommand{\Pc}{\mathcal{P}}

\definecolor{myblue}{rgb}{0.3328, 0.3539, 0.7758}
\definecolor{myblue2}{rgb}{0.0328, 0.0539, 0.4758}
\definecolor{mygreen2}{rgb}{ 0.0328 0.4758 0.0539} 
\definecolor{mygreen3}{rgb}{ 0.0328 0.1758 0.0539} 
\definecolor{myred}{rgb}{0.4758, 0.0328, 0.0539}
\definecolor{myred2}{rgb}{0.75, 0.0328, 0.0539}

\usepackage{amsthm}
\newtheorem{thm}{\bf{Theorem}}

\newtheorem{lem}{\bf{Lemma}}

\theoremstyle{remark} 

\newenvironment{theorem}
{\par\noindent \thm \begin{itshape}\noindent}
{\end{itshape}}
\newenvironment{lemma}
{\par\noindent  \lem \begin{itshape}\noindent}
{\end{itshape}}
\newcommand{\myCoLa}{
    \SetKw{KwEnd}{end}
	\textbf{Input}: Data matrix $\Amat$ distributed column-wise according to partition $\Pc$. Regularization parameter $\lambda$.
            
    \textbf{Initialize:} $\xhatinit=0\in\Rbb^{p\times 1}$ $\vvec_k^0{=}0\in\Rbb^{p\times 1}\forall\, k{=}1,\dots,K$ 
            
    \For{$t=0,\,1,\,\dots,\,T$}{
        $\vbar^t = \frac{1}{K}\sum_{k=1}^K\vvec_k^t$
        
        \For{$k \in \{1,\,2,\,\dots,\,K\}$}{
            $\cvec_k^t = \lambda\xhatt_k - \Amat_k\T(\yvec - \vbar^t)$
            
            $\dxt_k = -(K\Amat_k\T\Amat_k + \lambda\eye{p_k})\p\cvec_k^t$
            
            $\xhattp_k = \xhatt_k + \dxt_k$
            
            $\vvec_k^{t+1} = \vbar^t + K\Amat_k\dxt_k$
            
            \KwEnd  
        }{
        }
        \KwEnd
    }
        \caption{Implementation of \cocoa{} \cite{smith_cocoa_nodate} (and \cola{} \cite{he_cola_2019} with $\Wmat = \frac{1}{K} \bm{1}_K$) for  \eqref{eqn:problem} with \eqref{eqn:pinv_solver}.\label{alg:cola} }
        }

\begin{document}
\bstctlcite{IEEEexample:BSTcontrol}
\title{Generalization Error for Linear Regression under Distributed Learning\\
 }

\author{
    \IEEEauthorblockN{Martin Hellkvist, Ay\c ca \"Oz\c celikkale, Anders Ahl\'en  \thanks{M. Hellkvist and A.~\"Oz\c celikkale acknowledges the support from Swedish Research Council under grant 2015-04011.}}
    \IEEEauthorblockA{{Dept. of Electrical Engineering}, 
    {Uppsala University}, Sweden \\
    \{Martin.Hellkvist, Ayca.Ozcelikkale, Anders.Ahlen\}@angstrom.uu.se
    }
}

\maketitle

\begin{abstract}
Distributed learning facilitates the scaling-up of data processing by distributing the computational burden over several nodes.
Despite the vast interest in distributed learning,  generalization performance of such approaches is not well understood.
We address this gap by focusing on a linear regression setting.
We consider the setting where the unknowns are distributed over a network of nodes.
We present an analytical characterization of the dependence of the generalization error on the partitioning of the unknowns over nodes.
In particular, for the overparameterized case, our results show that while the error on training data remains in the same range as that of the centralized solution, the generalization error of the distributed solution increases dramatically compared to that of the centralized solution when the number of unknowns estimated at any node is close to the number of observations.
We further provide  numerical examples to verify our analytical expressions.
\end{abstract}


\section{Introduction} 
Distributed learning provides a framework for sharing the high computational burden of the learning task over multiple nodes, where the growing need for and interest from both academia and industry has led to a rapid advancement within the field \cite{verbraeken_survey_2019}.
Accordingly, distributed learning over wireless communication networks,
e.g., in the context of edge computing,
has emerged as a significant facilitator 
\cite{niknam_federated_2019, wang2020convergence}.
We contribute to the overall understanding of these methods by characterizing potential pitfalls of distributed learning for linear regression in terms of generalization error and by providing guidelines for best practice.

In a standard learning task, the main aim is to be able to estimate an observation $y$ when a corresponding input $\avec$ is given.  
Estimation of unknown model parameters using a set of training data, i.e., pairs of  $(y_i,\avec_i)$ is referred to as model training.
How well the trained model can explain the training data is referred to as the training error, i.e., the error that the model makes for the estimation of  $y_i$ in the training set. 
A key performance criterion for any trained model is the generalization error, i.e., how well a trained model can estimate a new observation $y$ given the corresponding $\avec$.
If the model performs well on new data, it is said to have low generalization error. 
In general, low training error does not always guarantee a low generalization error. Hence, it is of central interest to develop methods that have both low training and generalization error \cite{zhang_understanding_2017}. 
Modern machine learning techniques are often able to fit overparameterized models to exactly predict the training data, while still having low generalization error \cite{zhang_understanding_2017}.

Although various communications related challenges for distributed learning, such as energy constraints \cite{predd_distributed_2006}, quantization  \cite{magnusson_communication_2018}
and privacy \cite{niknam_federated_2019}, have been successfully investigated,
to the best of our knowledge there has been no attempt to characterize the generalization properties of distributed learning schemes.
In this article, we address this gap.
In contrast to the setting where the observations (for instance, sensor readings) are distributed over the nodes \cite{predd_distributed_2006},
our approach follows the line of work initiated by the seminal work of \cite{tsitsiklis_distributed_1984} where  the unknowns are distributed over the network.

We consider a linear model and  utilize the successful distributed learning method \cocoa{} \cite{smith_cocoa_nodate}. 
Our results show that the generalization performance of the distributed solution can heavily depend on the partitioning of the unknowns although the training error shows no such dependence, i.e., the distributed solution achieves training errors on the same level of accuracy as the centralized approach. 
Motivated by the success of overparameterized models in machine learning \cite{zhang_understanding_2017}  and recent results on the generalization error of such models \cite{belkin_reconciling_2019,belkin_two_2019}, we pay special attention to the overparameterized case, i.e., the number of unknowns is larger than the number of observations. 
In particular, if the number of unknowns assigned to any node is close to the number of observations, then the generalization error of the distributed solution may take extremely large values compared to the generalization error of the centralized solution.
Our main analytical results in Theorem~\ref{thm:x_tilde} and Lemma~\ref{lemma:x_tilde_t} present the expectation of the generalization error as a function of the partitioning of the unknowns. Furthermore, these analytical results are verified by  numerical results. 
Using these results, we provide guidelines for optimal partitioning of unknowns for distributed learning.

\textbf{Notation:}
    We denote the Moore-Penrose pseudoinverse and the transpose of a matrix $\Amat$ as $\Amat\p$ and $\Amat\T$, respectively.
    The $p\times p$ identity matrix is denoted as $\eye{p}$.
    We denote a column vector $\xvec \in \Rbb^{p \times 1}$ as $\xvec=[x_1; \, \cdots\, ; x_p]$, where the semicolon denotes row-wise separation. The matrix of all ones is denoted by $\bm{1}_K \in\Rbb^{K\times K}$.  
    Throughout the paper, we often partition matrices column-wise and vectors  row-wise. 
    Column-wise partioning of $\Amat \in \Rbb^{n\times p}$ into $K$ blocks with $\Amat_k\in\Rbb^{n\times p_k}$ is given by $\Amat= [\Amat_1, \, \cdots, \, \Amat_K]$.
    The row-wise partitioning of a vector $\xvec$ into $K$ blocks $\xvec_k\in\Rbb^{p_k\times 1}$ is given by $\xvec=[\xvec_1;\,\cdots;\,\xvec_K]$.

\section{Problem Statement}
    We focus on the linear model
        \begin{equation}\label{eqn:model}
            y_i = \avec_i\T \xvec+ w_i,
            \end{equation}
    where $y_i\in\Rbb$ is the $i$\textsuperscript{th} observation, $\avec_i\in\Rbb^{p\times 1}$ is the $i$\textsuperscript{th} regressor, $w_i$ is the unknown disturbance for the $i$\textsuperscript{th} observation, and $\xvec= [x_1;\,\cdots\,;x_p]\in\Rbb^{p\times1}$ is the vector of unknown coefficients.  
    
    We consider the problem of estimating $\xvec$ given $n$ data points, i.e., pairs of observations and regressors, $(y_i,\avec_i),~\irange,$ by minimizing the following regularized cost function: 
        \begin{equation}\label{eqn:problem}
            \min_{\xvec\in\Rbb^{p\times1}} \frac{1}{2}\norm{\yvec - \Amat\xvec}^2 + \frac{\lambda}{2} \norm{\xvec}^2,
        \end{equation}
    where $\Amat\in\Rbb^{n\times p}$ is the regressor matrix whose $i$\textsuperscript{th} row is given by $\avec_i^T\in\Rbb^{1\times p}$. 
    We further denote the first term as $f(\Amat\xvec) = \tfrac{1}{2}\norm{\yvec - \Amat\xvec}^2$.
    The second term  $\tfrac{\lambda}{2}\norm{\xvec}^2$ with $\lambda\geq 0$ denotes  the regularization function.

    We consider the setting where the regressors $\avec_i\T \in\Rbb^{1\times p} $ are independent and identically distributed (i.i.d.)  with $\avec_i \sim \Gauss{p}$. 
    Under this Gaussian regressor model, we focus on the generalization error of the solution to \eqref{eqn:problem} found by the distributed solver CoCoA \cite{smith_cocoa_nodate}.
    Our main focus is on the scenario where $\lambda = 0$, $w_i=0$ where the solutions with ${\lambda>0}$ are used for comparison.
    In the remainder of this section, we define the generalization error.
    We provide details about our implementation of CoCoA in Section~\ref{sec:dist}.
    
    Let $w_i=0$, $\forall i$, and let $\xhat$ be an estimate of $\xvec$ found by using the data pairs $(y_i, \avec_i),\,\irange$.
    For a given $\Amat$, the generalization error, i.e., the expected error for estimating $y $ when a new pair $(y,\avec)$ with $\avec \sim \Gauss{p}$ comes is given by 
    \begin{align}
        \Ebb_{a}[(y - \avec\T\xhat)^2 ] =& \Ebb_{a}[(\avec\T \xvec - \avec\T\xhat)^2] \\
        =& \Ebb_{a}[ \tr [  (\xvec-\xhat) (\xvec-\xhat)\T \avec \avec\T ]] \label{eqn:test:crossvanish} \\
        =& \| \xvec-\xhat\|^2, \label{eqn:test:avanish}
    \end{align}
     where $\avec$ is statistically independent of $\Amat$ and we have used the notation $\Ebb_{a}[\cdot]$ to emphasize that the expectation is over $\avec$.
     Here \eqref{eqn:test:avanish} follows from  $\avec \sim \Gauss{p}$.
     We are interested in the expected generalization error over the distribution of training data
        \begin{align}\label{eqn:generalization_error_def}
            \epsilon_G =& \Ebb_{\Amat} [ \| \xvec-\xhat\|^2],
        \end{align}
    where the expectation is over the regressor matrix $\Amat$  in the training data.
    In the rest of the paper, we focus on the evolution of $\epsilon_G$ in  CoCoA.
    For notational simplicity, we drop the subscript $\Amat$ from our expectation expressions.
    \kern-0.1em
\section{Distributed Solution Approach}\label{sec:dist}
  \kern-0.1em

    \SetAlgoSkip{bigskip}
    \setlength{\textfloatsep}{0pt}
    \begin{algorithm}[t]
        \myCoLa
    \end{algorithm}
    
    As the distributed solution approach, we use the iterative approach \cocoa{} introduced in \cite{smith_cocoa_nodate}. In \cocoa{}, mutually exclusive subsets of coefficients of $\xvec$ and the associated subset of columns of $\Amat$ are distributed over $K$ nodes ($K \leq p$).  
    Hence, the $p$ unknown coefficients are partitioned over $K$ nodes so that each node governs the learning of $p_k$ variables, hence $\sum_{k=1}^K p_k =p$.
    We denote the part of $\Amat$ available at node $k$ as $\Amat_k\in\Rbb^{n\times p_k}$.
    In particular, using this partitioning, $\yvec$ with $w_i=0$, $\forall i$, can be expressed as 
    \begin{align}
     \yvec= \Amat \xvec = [\Amat_1,\cdots,\Amat_K]         \begin{bmatrix}
            \xvec_1\\\vdots\\\xvec_K
        \end{bmatrix} =\sum_{k=1}^K \Amat_k \xvec_k,
    \end{align}
    where $\xvec_k$ is the partition at node $k$.
    Note that there is no loss of generality due to the specific order of this partitioning structure since the columns of $\Amat$ are i.i.d. (since rows are i.i.d. with $\Gauss{p}$).
    
    In \cocoa{}, at iteration $t$, node $k$ shares its estimate of $\yvec$, denoted $\vvec_k^{t}$, over the network.
    Note that the $\Amat_k$'s and the observation vector $\yvec$ are fixed over all iterations. 
    The variables $\xhatt_k\in\Rbb^{p_k\times 1}$ and $\dx^t_k\in\Rbb^{p_k\times 1}$ are the estimate and its update computed by node $k$, respectively.
    Hence, $\xhatt$ and $\dx^t$ are partitioned as $\xhatt = [\xhatt_1; \cdots; \xhatt_K]$ and $\dx^t=[\dx^t_1;\cdots;\dx^t_K]$.
    The average over all local estimates $\vvec_k^t$ is denoted as $\vbar^t$.
    
    At iteration $t$, \cocoa{} solves the following minimization problem at each node \cite{smith_cocoa_nodate}:
        \begin{align}\label{eqn:problem_t}
        \begin{split}
            \min_{\dx^t_k} \nabla_{\vbar^t} f(\vbar^t)\T\Amat_k\dx^t_k &\\ 
            + \tfrac{\sigma'}{2\tau}\norm{\Amat_k\dx^t_k}^2&+\tfrac{\lambda}{2}\norm{\xhatt_k+\dx^t_k}^2.
        \end{split}
        \end{align}

    Using $f(\Amat\xvec) = \tfrac{1}{2}\norm{\yvec - \Amat\xvec}^2$, we have the smoothness parameter $\tau=1$\cite{he_cola_2019}.
    We set $\sigma'=K$ since it is considered a safe choice\cite{he_cola_2019}.
    Only keeping the terms that depend on $\dx^t_k$ reveals that the solution to \eqref{eqn:problem_t} can be equivalently found by solving the following problem
        \begin{align}\begin{split}\label{eqn:problem_open}
            \min_{\dx^t_k} & ~ (\dx^t_k)\T( \tfrac{K}{2}\Amat_k\T\Amat_k + \tfrac{\lambda}{2} \eye{p_k})\dx^t_k\\
            & + ( \lambda\xhatt_k - \Amat_k\T(\yvec - \vbar^t) )\T\dx^t_k.
        \end{split}
        \end{align}
    Taking the derivative with respect to $\dx^t_k$ and setting it to zero, we obtain
        \begin{align}\begin{split}\label{eqn:iteration_eqn_system}
            (K & \Amat_k\T\Amat_k + \lambda \eye{p_k})\dx^t_k = - ( \lambda\xhatt_k - \Amat_k\T(\yvec - \vbar^t) ).
        \end{split}\end{align}
    With $\lambda=0$, existence of a matrix inverse is not guaranteed.   
    Hence, the local solvers use Moore-Penrose pseudoinverse  to  solve  \eqref{eqn:iteration_eqn_system} as 
        \begin{equation}\label{eqn:pinv_solver}
            \dx^t_k = - (K\Amat_k\T\Amat_k + \lambda \eye{p_k})\p ( \lambda\xhatt_k - \Amat_k\T(\yvec - \vbar^t) ).
        \end{equation}
    The resulting algorithm for estimating $\xvec$ iteratively is presented in Algorithm~\ref{alg:cola}.

    In \cite{he_cola_2019}, a generalization of \cocoa{} is presented, named \cola{}, where a mixing matrix $\Wmat$ is introduced to model the quality of the connection between nodes.
    For $\Wmat=\tfrac{1}{K}\bm{1}_K$ \cola{} reduces to \cocoa{}, hence our analysis also applies to this special case of \cola{}.
\section{Partitioning and the Generalization Error}\label{sec:theorems}
   This section presents our main results in Theorem \ref{thm:x_tilde} and Lemma~\ref{lemma:x_tilde_t}, which reveal how the generalization error changes based on the data partitioning.
   We first provide a preliminary result to describe the evolution of the estimates of Algorithm~\ref{alg:cola}:
        \begin{lemma}\label{lemma:update-eqn}
    Using Algorithm \ref{alg:cola} with $\lambda=0$,  the closed form expression for $\xhattp$ is given by
    
    \begin{align} 
        \xhattp = \Big(
        \eye{p} - \frac{1}{K}   
        \begin{bmatrix}
            \Amat_1\p\\\vdots\\\Amat_K^+
        \end{bmatrix}
        \Amat\Big) \xhatt + \frac{1}{K} 
        \begin{bmatrix}
            \Amat_1\p\\\vdots\\\Amat_K^+
        \end{bmatrix}
        \yvec.
    \end{align}
\end{lemma}
        
    Proof: See Section~\ref{sec:pf:lemma:update-eqn}.
    This result shows that when $\lambda=0$, the estimate in each iteration is a combination of the previous global estimate ($\xhatt$) and the local least-squares solutions ($\Amat_k\p \yvec$) from each node.
    We now present our main results: 
        
        \begin{theorem}\label{thm:x_tilde}
            Let $\Amat\in\Rbb^{n\times p}$ have i.i.d. rows with ${\avec_i \sim \Gauss{p}}$.
Using Algorithm\,\ref{alg:cola} with ${\lambda{=}0}$, ${w_i{=}0},\,\forall i$, the generalization error in iteration $t=1$, i.e.,  $\epsilon_G$, is given by
        \begin{equation}\label{eqn:xtildetp}\textstyle
            \Ebb\left[\norm{\xvec - \xhat^1}^2\right]=\sum_{k=1}^K \norm{\xvec_k }^2 \alpha_k,
        \end{equation}
    where $\alpha_k$ and $\gamma_k$, $ \,k=1,\,\dots,K$, are given by
        \begin{equation}\textstyle\label{eqn:alpha_k}
            \hspace{-47pt}\alpha_k = \frac{1}{K^2}(K^2 + (1-2K)\tfrac{r_{\min,k}}{p_k} + \sum_{\substack{i=1\\i\neq k}}^K\gamma_i),
        \end{equation}
        \begin{subnumcases}{\label{eqn:gamma_main}\hspace{-10pt} \gamma_k=}
            \tfrac{r_{\min, k}}{r_{\max, k} - r_{\min, k} - 1} & \hspace{-15pt}for $p_k \notin \{n-1, n, n+1\},$ \label{eqn:gamma_main_a}\\
            +\infty & \hspace{-15pt}otherwise, \label{eqn:gamma_main_b}
        \end{subnumcases}
and $r_{\min, k}=\min\{p_k,n\}$ and $r_{\max, k}=\max\{p_k,n\}$.
\vspace{3pt}
        \end{theorem}
        
    Proof: See Section \ref{sec:pf:thm:x_tilde}.
        Here, while writing the expressions, we have used the notational convention that if any $\alpha_k=+\infty$ and the corresponding $\norm{\xvec_k}^2 = 0$, then that component of \eqref{eqn:xtildetp} is also zero.  
    Note that the infinity, i.e., $\infty$, in \eqref{eqn:gamma_main_b} denotes the indeterminate/infinite values due to divergence of the relevant integrals. 
    \iftr
    Further discussions on this point are provided together with an illustrative example in  Section~\ref{sec:example1by2}. 
    \else
    Further discussions on this point are provided together with an illustrative example in  \cite{HellkvistOzcelikkaleAhlen_distributed2020_technicalReport}.
    \fi

    Theorem~\ref{thm:x_tilde} shows how the partitioning of $\xvec$ (and hence $\Amat$) over the nodes affects the generalization error $\epsilon_G$.
    Note that the interesting case of $p_k\in\{n-1,n,n+1\}$, $K>1$, occurs with the overparameterized scenario of $n\leq p$. 
    If $\xvec_k\neq\bm{0}$ and any ${p_i\in\{n-1,n,n+1\}},\,i\neq k$, the generalization error after the first iteration will be extremely large, since the corresponding $\alpha_k$ in \eqref{eqn:xtildetp} will be extremely large.
    In order to avoid large generalization errors, no partition $\Amat_k$ should have a number of columns $p_k$ close to the number of observations $n$.
    Note that according to \eqref{eqn:alpha_k}, having $p_k\in\{n-1,n,n+1\}$ in one node affects the generalization error associated with the partition in the other nodes.

    We now consider evolution of the generalization error: 
        \begin{lemma}\label{lemma:x_tilde_t}
            Consider the setting in Theorem \ref{thm:x_tilde}. For large $t$, the generalization error associated with $\xhattp$  is given by
    \begin{equation}\label{eqn:gen:approx:xhattp}\textstyle
        \Ebb[\norm{\xvec - \xhattp}^2] \approx \sum_{k=1}^K  \alpha_k \Ebb[\norm{\xvec_k - \xhatt_k}^2] 
    \end{equation}
where $\alpha_k$ is defined as in \eqref{eqn:alpha_k}.
\vspace{3pt}
        \end{lemma}
        
    \iftr
    Proof: See Section~\ref{sec:pf:thm:x_tilde_t}. 
    \else
    Due to page limitations, the proof is presented in \cite{HellkvistOzcelikkaleAhlen_distributed2020_technicalReport}.
    \fi
    Lemma \ref{lemma:x_tilde_t} reveals that if we have $\Ebb[\norm{\xvec_k - \xhat_k^t}^2]\neq 0$ with  ${p_i\in\{n-1,n,n+1\}},\,i\neq k,$ at a given iteration, then the average generalization error will increase dramatically in the next iteration.
    Hence, if the average generalization error takes large values, it will not decrease by iterating the algorithm further. Numerical illustrations are provided in Section~\ref{section:numerical}. 
    
    Following \cite{breiman_how_nodate}, a similar analysis is presented  in \cite{belkin_two_2019}, to explain the ``double descent'' curves in \cite{belkin_reconciling_2019}.
    The analyses of \cite{belkin_two_2019,breiman_how_nodate}  focus on the centralized problem where only a subset $\bar{p}$ of the $p$ unknowns are learnt and present how the generalization error increases when $\bar{p}$ is close to the number of observations $n$. 
    In this paper we extend these results for distributed learning with \cocoa{}.

    We note that presence of noise $w_i$ in \eqref{eqn:model} during training would provide some numerical stability.
    Similarly, having a non-zero regularization during training,~i.e., $\lambda>0$, will make the matrix in \eqref{eqn:pinv_solver} invertible, hence replacing the pseudo-inverse of \eqref{eqn:pinv_solver} with an inverse.
    With a large enough $\lambda>0$ (compared to the machine precision), this will provide numerical stability which can reduce the large values in the generalization error significantly, at the cost of a larger training error.
    On the other hand, a too large regularization will make the distributed solution penalize the norm of the solution too much, and the solution will neither fit the training data nor the test data.
    We illustrate these effects in Section~\ref{section:numerical}.

\section{Numerical Examples}\label{section:numerical}
    
    We now provide numerical results to illustrate the dependence of the generalization error on the partitioning and the effect of regularization. 
    We generate $\xvec$ with $\xvec\sim\Gauss{p}$ once in the numerical experiments and keep it fixed. We generate the rows of $\Amat\in\Rbb^{n\times p}$ i.i.d. with distribution $\Gauss{p}$. We set $n=50$, $p=150$,  $w_i=0,\,\forall i$. The data is partitioned over $K=2$ nodes, so $p = p_1+p_2$.
    
\begin{figure}
            \centering
            \includegraphics[width=0.49\textwidth, trim=0.1cm 0.5cm 0cm .36cm,clip]{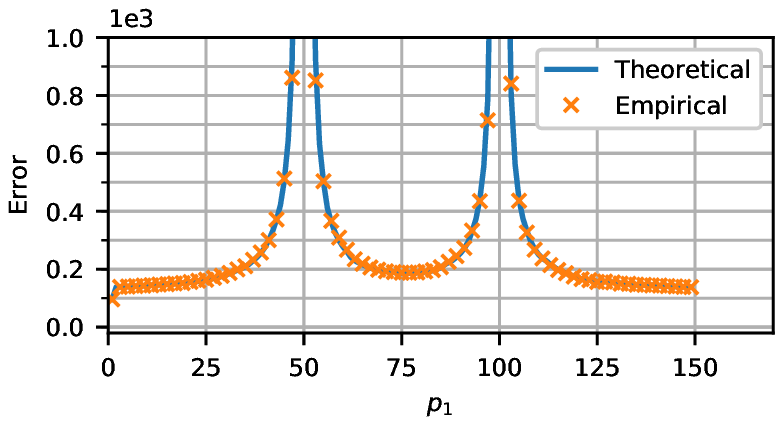}
             \caption{Comparison of  $\Ebb[\norm{\xvec - \xhat^1}^2]$  expression in \eqref{eqn:xtildetp} with the empirical ensemble average for $K=2$,  $\lambda=0$.}
            \label{fig:expectation-sweep-fig}
        \end{figure}

    {\it{Verification of  Theorem \ref{thm:x_tilde}}}: We first empirically verify the expression in \eqref{eqn:xtildetp} from Theorem \ref{thm:x_tilde}.
    We obtain the empirical results  by computing the first iteration of Algorithm \ref{alg:cola} for $N=100$ simulations.
    Note that these values correspond to the average generalization error (i.e. risk) by \eqref{eqn:generalization_error_def}. 
    In Figure \ref{fig:expectation-sweep-fig}, we present the analytical value $\epsilon_G$, ~i.e, ${\Ebb[\norm{\xvec-\xhat^1}^2]}$ from~\eqref{eqn:xtildetp}, and the empirical average ${\tfrac{1}{N}\sum_{i=1}^N \norm{\xvec-\xhat^1_{(i)}}^2}$, where the subscript $(i)$ denotes the $i$\textsuperscript{th} simulation.
    Figure \ref{fig:expectation-sweep-fig} illustrates that the empirical average follows the analytical values for all $p_1$ in \eqref{eqn:gamma_main}.
    When  $p_k\in\{n-1,n,n+1\}$, the empirical average increases so drastically the values are out of the range of the plots.
    For $p_k \approx n$, $p_k \notin \{n-1,n,n+1\} $, we see that empirical values take large values and these values are exactly on the analytical curve. 
    Note that no analytical value is computed for $p_k\in\{n-1,n,n+1\}$, hence the increase in the analytical expressions around $p_k \approx n$, $p_k \notin \{n-1,n,n+1\} $ directly comes from large but finite values dictated by the analytical expression. 

    {\it{Generalization error after convergence}}:  We now illustrate that the generalization error does not decrease when Algorithm~\ref{alg:cola} is run until convergence.
    We set the number of iterations for Algorithm~\ref{alg:cola} as $T=200$. 
    We note that increasing $T$ further does not change the nature of the results.
    The average training error is calculated as ${\frac{1}{Nn}\sum_{i=1}^N\norm{\Amat_{(i)}(\xvec - \xhat_{(i)}^T)}^2}$, where the superscript $T$ denotes the final iteration.
    The generalization error is calculated in a similar fashion but using a new data matrix $\Amat'\in\Rbb^{10n\times p}$ from the same distribution as $\Amat\in\Rbb^{n\times p}$.
    Here $\Amat$, $\Amat'$ are independently sampled for each simulation.
    The matrix $\Amat'$ is chosen to have $10n$ rows so that the generalization error is averaged over a large number of data points.
    For benchmarking, we use the training and the generalization errors of the centralized least-squares (LS) solution, i.e., $\xhat = \Amat\p \yvec$ using the whole $\Amat$.

        \begin{figure}
            \centering
            \includegraphics[width=0.49\textwidth, trim=0.1cm 0.36cm 0cm .36cm,clip]{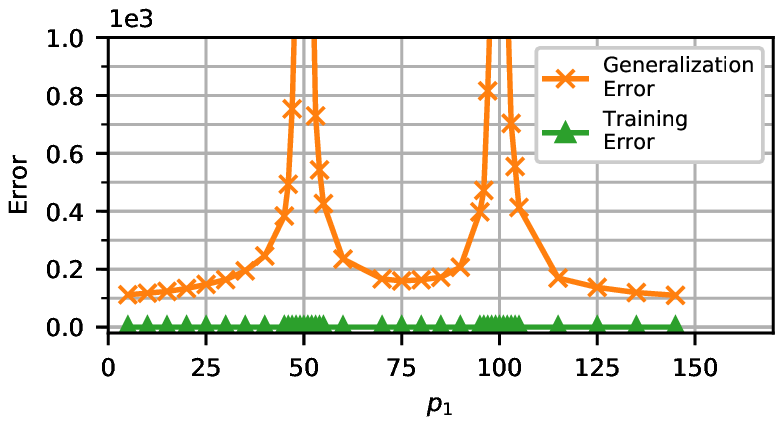}
            \caption{The generalization error and the training error for ${K=2}$, $\lambda=0$ after convergence.}
            \label{fig:main-fig}
        \end{figure}

    In Figure \ref{fig:main-fig}, we plot the empirical average of the generalization error and the training error of Algorithm~\ref{alg:cola} as a function of $p_1$ with $\lambda=0$. 
    When either $p_1$ or $p_2$ approaches $n=50$, there is a large increase in the generalization error.
    This behaviour is consistent with the general trend of Figure \ref{fig:expectation-sweep-fig}, which was obtained using Theorem~\ref{thm:x_tilde}.
    This numerical result supports the result of Lemma~\ref{lemma:x_tilde_t}, i.e., once the generalization error increases drastically,  iterations of Algorithm \ref{alg:cola} do not decrease it.
    In particular, the peak generalization error for Algorithm~\ref{alg:cola} is on the order of $10^5$ (not shown on the plot).
    On the other hand, the distributed solution fits the training data perfectly, as does the LS solution: the respective  training errors are lower than $10^{-25}$.
    In contrast to the distributed case, the LS solution fits the new data well with an average generalization error of $\approx 60$.

    Hence, although Algorithm~\ref{alg:cola} successfully finds a solution that achieves a training error on the same level with the direct centralized solution, the generalization error is significantly higher when $p_k\in\{n-1,n,n+1\}$.
    
    \textit{Effect of regularization:} We now investigate the effects of regularization on the peaks of Figure \ref{fig:main-fig}.
    We set a non-zero regularization parameter $\lambda$ and run the same simulations as in Figure \ref{fig:main-fig}.
    A   value of $\lambda$ between $10^{-4}$ and $10^3$ dampens the peaks in generalization error (when $p_1$ is close to $50,\,100$) to between $10^4$ and $10^2$.
    As $\lambda$ is increased beyond $10^{-4}$, the training error starts to grow.
    In particular, for $\lambda=10^3$, the training error is on the same level as the generalization error.
    Any further increase in $\lambda$ increases both the training and the generalization error.
    These results are consistent with the discussions in Section~\ref{sec:theorems}.
    
\kern-0.2em    
\section{Conclusions}
\kern-0.2em  
    We have presented a characterization of the generalization error showing how partitioning  plays a major role in  distributed linear learning.
    In particular, our analytical results show how it is crucial for the generalization performance that the partitioning must avoid setting the number of unknowns in any node  close to the number of available observations.
    We have presented numerical results, simulating the distributed learning system \cocoa{}, verifying our analytical results. Extension of this work to the fully decentralized case of \cola{} is considered an important direction for future work. 
    
\kern-0.1em
\section{Appendix}

\subsection{Proof of Lemma \ref{lemma:update-eqn}}\label{sec:pf:lemma:update-eqn}
    
        For $\lambda=0$, the formula for $\dx_k^t$ reduces to 
        \begin{equation}\label{eqn:reduced_dxk}
            \dx_k^t = \tfrac{1}{K}\Amat_k\p(\yvec - \vbar^t).
        \end{equation}
        Using
        $\xhat^0 = \bm{0}$, $\xhat_k^{t} = \xhat_k^{t-1} + \dx_k^{t-1}$
        and the partitioning structure  
        $\dx=[\dx_1;\cdots;\dx_K]$, we obtain 
        $\xhatt = \sum_{i=0}^{t-1}\dx^i$.
        Together with
        $\vvec_k^{t} 
        = \vbar^{t-1} + K\Amat_k\dx_k^{t-1} $
        and $\Amat=[\Amat_1,\cdots,\Amat_K]$ 
        we have
            \begin{align}\textstyle
            \begin{split}
                \vbar^t &= \frac{1}{K}\sum_{k=1}^K \vvec_k^{t} 
                = \vbar^{t-1} + \sum_{k=1}^K \Amat_k\dx_k^{t-1} \\
                &= \vbar^{t-1} + \Amat\dx^{t-1}
                =\vbar^0 + 
                \sum_{i=0}^{t-1}\Amat\dx^{i} 
                = \Amat\xhatt.
                \label{eqn:lambda=0_vtp1}
            \end{split}
            \end{align}
        Combining \eqref{eqn:reduced_dxk} and \eqref{eqn:lambda=0_vtp1} with $\xhattp_k = \xhatt_k + \dx_k^t$, we obtain
            \begin{equation}
                \xhattp_k = \xhatt_k - \tfrac{1}{K}\Amat_k\p \Amat\xhatt + \tfrac{1}{K}\Amat_k\p\yvec.
            \end{equation}
            Putting this result in  vector form gives the desired expression. 

\subsection{Proof of Theorem \ref{thm:x_tilde}}\label{sec:pf:thm:x_tilde}
    
    Let us define the matrix consisting of pseudo-inverses of blocks of ${\Amat}$  as follows:
    \begin{align}\label{eqn:def:Amat}
     \Bar{\Amat}= \Aplusmathorz \in \Rbb^{p \times n}
    \end{align}
    Now we consider the error for the unknown $\xvec$ at iteration $t=1$, i.e.  $\xtilde^1 = \xvec - \xhat^{1}$.
    Using Lemma~\ref{lemma:update-eqn} and the fact that 
    $\yvec = \Amat\xvec$, 
    we have an expression for $\xtilde^1$ as follows 
    ${\xtilde^1 = \left(\eye{p} - \tfrac{1}{K}\Bar{\Amat} \Amat \right) \, \xtilde^0= \left(\eye{p} - \tfrac{1}{K}\Bar{\Amat} \Amat \right) \, \xvec}$
    where we have used that $\xtilde^0 = \xvec - \xhat^0 = \xvec$ since the algorithm is initialized with $\xhat^0=\bm{0}$. 
    Hence, the error after one iteration is expressed in terms $\xvec$. 
    We now consider the expectation of $\norm{\xtilde^1}^2$, i.e.  
    \begin{align}
        \Ebb[\norm{\xtilde^1}^2] &= \Ebb[\norm{(\eye{p} - \tfrac{1}{K}\Bar{\Amat} \Amat)\xvec}^2] \\
        \label{eqn:norm_2}
        &=\xvec\T\Ebb[(\eye{p}-\tfrac{1}{K}\Qmat\T)(\eye{p}-\tfrac{1}{K}\Qmat)]\xvec\\
        \label{eqn:xtilde_norms}
        &= \norm{\xvec}^2 - \tfrac{2}{K} \xvec\T\Ebb[\Qmat]\xvec + \tfrac{1}{K^2} \xvec\T\Ebb[\Qmat\T\Qmat]\xvec 
    \end{align}
    where in \eqref{eqn:norm_2}, we have introduced the notation $\Qmat = \Bar{\Amat} \Amat$.
    
    In \eqref{eqn:xtilde_norms}, we will first evaluate the term $\xvec\T\Ebb[\Qmat]\xvec$, then
    $\xvec\T\Ebb[\Qmat\T\Qmat]\xvec$ and finally combine these results to find an expression for  $\Ebb[\norm{\xtilde^1}^2]$.
    
    We now evaluate the term $\xvec\T\Ebb[\Qmat]\xvec$.
    The matrix $\Qmat$ can be expressed as follows:
        \begin{equation}
            \Qmat = \Bar{\Amat}\Amat = \Aplusmat\Amat = 
            \begin{bmatrix}
                \Amat_1\p\Amat_1 & \cdots & \Amat_1\p\Amat_K \\
                \vdots           & \ddots & \vdots \\
                \Amat_K\p\Amat_1 & \cdots & \Amat_K\p\Amat_K
            \end{bmatrix}.
        \end{equation}
    Since $\Amat_k$ and $\Amat_i$ are statistically independent for $k\neq i$, and  $\Ebb[\Amat]=\bm{0}$, we obtain
        \begin{equation}
            \Ebb[\Qmat] = 
            \begin{bmatrix}
                \Ebb[\Amat_1\p\Amat_1] & \cdots & \bm{0} \\
                \vdots           & \ddots & \vdots \\
                \bm{0} & \cdots & \Ebb[\Amat_K\p\Amat_K]
            \end{bmatrix}.
        \end{equation}
        The quadratic form $\xvec\T\Ebb[\Qmat]\xvec$ can then be expressed as the following summation:
        \begin{equation}\textstyle\label{eqn:fullstop_QMat}
            \xvec\T\Ebb[\Qmat]\xvec = \sum_{k=1}^K \xvec_k\T\Ebb[\Amat_k\p\Amat_k]\xvec_k.
        \end{equation}
       As an intermediate step, we now present Lemma \ref{lem:wisharts-nice-dream} which will be utilized throughout the proofs: 
            \begin{lemma}\label{lem:wisharts-nice-dream}
        Let $\Cmat\in\Rbb^{n \times p_c}$ be a random matrix with i.i.d. rows with the distribution $\Gauss{p_c}$.
        Let $\zvec\in\Rbb^{p_c\times 1}$.
        Then 
            \begin{align}
                \zvec\T\Ebb[\Cmat\p\Cmat]\zvec = \norm{\zvec}^2\,\frac{\bar{r}_{\min}}{p_c},
            \end{align}
            where $\bar{r}_{\min}=\min\{n,p_c\}$.
           \vspace{3pt}
\end{lemma}
        
        \iftr
            Proof: See Section~\ref{sec:pf:lem:wisharts-nice-dream}.
        \else
            Proof: See \cite{HellkvistOzcelikkaleAhlen_distributed2020_technicalReport}.
        \fi
     This type of expressions have been utilized before, e.g., in \cite{belkin_two_2019} for $n \leq p_c$  without a proof. This result follows from the unitary invariance  (and invertibility of square) Gaussian matrices. 
     \iftr
        We provide a proof for the sake of completeness in
        Section~\ref{sec:pf:lem:wisharts-nice-dream}.
    \else
        We provide a proof for the sake of completeness in \cite{HellkvistOzcelikkaleAhlen_distributed2020_technicalReport}. 
    \fi
        
    By definition of $\Amat$, the rows of $\Amat$ are i.i.d. with $\Gauss{p}$.
    Hence, the rows of $\Amat_k$ are i.i.d. with $\Gauss{p_k}$, for any $k$. 
    Thus,  using Lemma \ref{lem:wisharts-nice-dream} with $\Cmat= \Amat_k$ in \eqref{eqn:fullstop_QMat}, we obtain
        \begin{equation}\label{eqn:tidepod_xtilde}
            \xvec\T\Ebb[\Qmat]\xvec = \sum_{k=1}^K \norm{\xvec_k}^2 \frac{r_{\min,k}}{p_k}.
        \end{equation}

    We now consider the term $\xvec\T\Ebb[\Qmat\T\Qmat]\xvec$ in Lemma \ref{lem:A^+A}:
  
    \begin{lemma}\label{lem:A^+A}
    Let $\Amat$ be a $n\times p$ random matrix with i.i.d. rows with the distribution $\Gauss{p}$.
    Let $\Bar{\Amat}$ denote the matrix $[\Amat_1\p;\cdots;\,\Amat_K\p] \in \Rbb^{p \times n} $.
    Let $\zvec=[\zvec_1;\,\cdots;\,\zvec_K]\in\Rbb^{p\times 1}$, where $\zvec_k\in\Rbb^{p_k\times 1}$ and $r_{\min,k} = \min\{n, p_k\},~r_{\max,k} = \max\{n, p_k\},~k=1,\,\dots,\,K$.
    Then
        \begin{equation}\textstyle
            \kern -.4em \zvec\T\Ebb[\Amat\T\Bar{\Amat}\T\Bar{\Amat}\Amat]\zvec {=} \sum_{k=1}^K \norm{\zvec_k}^2\left(\frac{r_{\min,k}}{p_k} {+} \sum_{\substack{i=1\\i\neq k}}^K \gamma_i\right),
        \end{equation}
    with $\gamma_k,\,k=1,\,\dots,\,K,$ defined in \eqref{eqn:gamma_main}.
   \vspace{3pt}
\end{lemma}
    
    \iftr    
        Proof: See Section~\ref{sec:pf:lem:A^+A}.
    \else
        Proof: See \cite{HellkvistOzcelikkaleAhlen_distributed2020_technicalReport}.
    \fi
    Using $\Qmat = \Bar{\Amat}\Amat$ and Lemma~\ref{lem:A^+A}, we obtain
        \begin{equation}\label{eqn:quadform_thm}\textstyle
            \xvec\T\Ebb[\Qmat\T\Qmat]\xvec {=} \sum_{k=1}^K\norm{\xvec_k}^2\left(\frac{r_{\min,k}}{p_k} {+} \sum_{\substack{i=1\\i\neq k}}^K \gamma_i\right).
        \end{equation}

    Combining \eqref{eqn:quadform_thm} and \eqref{eqn:tidepod_xtilde} with \eqref{eqn:xtilde_norms}, we obtain  \eqref{eqn:xtildetp} of Theorem~\ref{thm:x_tilde}.
        This concludes the proof 
        of Theorem~\ref{thm:x_tilde}.

\iftr
\subsection{An Illustrative Example}\label{sec:example1by2}
We now consider the special case where $n=1,\,p=2,\,K=2$, $p_1=1$, $p_2=1$ where $\Amat= [a_1, a_2] \in \Rbb^{1 \times 2}$,  $\xvec= [x_1; x_2] \in \Rbb^{2 \times 1}$. Hence, $y= a_1 x_1+ a_2 x_2 $. Now consider the case where $a_1$ and $a_2$ are non-zero so that the pseudo-inverses are given by $\frac{1}{a_1}$ and $\frac{1}{a_2}$, respectively.
Note that this is the case with probability one since $a_1$ and $a_2$ are Gaussian distributed. 
By Lemma~\ref{lemma:update-eqn} and $\xhat^0=0$, we have $\hat{x}_1^1 =  \frac{1}{2 a_1} y = \frac{1}{2} x_1 +\frac{a_2}{2 a_1} x_2 $ and $\hat{x}
_2^1= \frac{1}{2 a_2} y= \frac{a_1}{2 a_2} x_1+ \frac{1}{2} x_2$. Hence, 
\begin{align}
    \nonumber
    \Ebb[\| \xvec -\xhat^{1}\|^2_2] 
    &= \Ebb\Big[| \frac{x_1}{2} -\frac{a_2}{2 a_1} x_2|^2  +
     |\frac{x_2}{2} 
     -\frac{a_1}{2a_2} x_1 |^2\Big]  \\
     \nonumber
    &= \Ebb\Big[|\frac{a_2}{2 a_1}x_2|^2 
    +
    | \frac{x_1}{2}|^2
    -\frac{a_2}{ 2 a_1} x_1 x_2 ] \\
    & \quad 
    + | \frac{a_1}{2 a_2} x_1|^2 +
    |\frac{x_2}{2}|^2 
    - \frac{a_1}{ 2 a_2} x_1 x_2 \Big]
\end{align}
Now consider the individual terms, for instance  $\Ebb[|\frac{a_2}{2 a_1}x_2|^2 ] = \Ebb[a_2^2] \Ebb[\frac{1}{ a_1^2}] \frac{x_2^2}{4}$, 
where we have used the statistical independence of $a_1$ and $a_2$.  Here, $\Ebb[a_2^2]$  is finite valued. On the other hand,  for  $a_i$ Gaussian distributed,  $\Ebb[\frac{1}{a_i^2}]$ diverges (and also note that  $\int_{\epsilon}^{\infty} \frac{1}{a_i^2} \exp(-a_i^2) \,da_i$  takes large values for any given finite $\epsilon>0$). Similar conclusions can be drawn for the other terms in the expectation. Hence, these observations illustrate the infinite/indeterminate values in Theorem~\ref{thm:x_tilde}. 

On the other hand, when at least one of the $a_i$'s is zero (note that this event has probability zero), the associated pseudo-inverse is zero.  By straightforward calculations, the average generalization error can be found to be finite in this case.
\fi

\iftr
    \subsection{Proof of Lemma \ref{lemma:x_tilde_t}}\label{sec:pf:thm:x_tilde_t}
    
    We adopt the same notation in the proof of Theorem \ref{thm:x_tilde} in Section~\ref{sec:pf:thm:x_tilde}.  We consider the error in the estimate $\xhattp$, i.e.  $\xtilde^{t+1} = \xvec - \xhattp$ for an arbitrary $t$.
    Using Lemma~\ref{lemma:update-eqn} and 
    $\yvec = \Amat\xvec$, $\xtilde^t$ can be written as
    ${\xtildetp = \left(\eye{p} - \tfrac{1}{K}\Bar{\Amat} \Amat \right) \xtildet}$.
    Now, consider the evolution of the expected error, i.e.,
    \begin{align}
        &\Ebb[\norm{\xtildetp}^2] = \Ebb[\norm{(\eye{p} - \tfrac{1}{K}\Bar{\Amat} \Amat)(\xtildet)}^2] \\
        \label{eqn:norm_2:app}
        &=\Ebb[(\xtildet)\T(\eye{p}-\tfrac{1}{K}\Qmat\T)(\eye{p}-\tfrac{1}{K}\Qmat)\xtildet]\\
        \label{eqn:longsum_xtilde:app}
        &\approx  \Ebb[ \norm{\xtildet}^2] - \tfrac{2}{K} \Ebb[(\xtildet)\T \Ebb[\Qmat] \xtildet]\\
        \nonumber
        &\quad\, + \tfrac{1}{K^2} \Ebb[ (\xtildet)\T \Ebb[ \Qmat\T \Qmat] \xtildet ]         
    \end{align}
    where $\Qmat = \Bar{\Amat} \Amat$. 
    In \eqref{eqn:longsum_xtilde:app}, we have used the Independence Assumption \cite[Ch.16]{b_HaykinAdaptive}, which  assumes statistical independence between $\xhatt$ and the regressors in $\Amat$ for large $t$.
    
    Independence Assumption \cite[Ch.16]{b_HaykinAdaptive} is a widely utilized assumption in signal processing literature to study the transient and the steady-state behaviour of adaptive filters yielding extraordinary agreement between analytical studies and empirical values, see for instance \cite[Ch.16.6]{b_HaykinAdaptive}.
    In our particular case, the assumption does not hold to its full extent since we have an overparametrized system of equations with multiple solutions. 
    Instead, numerical studies suggest that for large $t$ there may be a constant finite gap between the actual values and the approximation, for instance in between $\Ebb[(\xtildet)\T \Qmat\T \Qmat \xtildet]$ and $\Ebb[(\xtildet)\T \Ebb[\Qmat\T \Qmat] \xtildet]$.  
    Note that since this gap is finite and constant, the generalization error will still grow whenever  $\gamma_i=+\infty$.
    
    Now following the proof of Theorem \ref{thm:x_tilde} with \eqref{eqn:longsum_xtilde:app} instead of \eqref{eqn:xtilde_norms}, we obtain the expression  in \eqref{eqn:gen:approx:xhattp} of Lemma~\ref{lemma:x_tilde_t}.
    This concludes the proof of Lemma~\ref{lemma:x_tilde_t}.
\fi

\iftr
    \subsection{Proof of Lemma \ref{lem:wisharts-nice-dream}}\label{sec:pf:lem:wisharts-nice-dream}   
        Let us denote the singular value decomposition of $\Cmat\in\Rbb^{n\times p_c}$ as 
        \begin{equation}
            \Cmat = \Umat\T\Lambmat\Vmat,
        \end{equation}
    where $\Umat \in \Rbb^{n\times n}$ and $\Vmat \in \Rbb^{p_c\times p_c}$ are unitary matrices (which reduces to real orthonormal matrices  since $\Cmat$ is real-valued) and $\Lambmat \in \Rbb^{n\times p_c}$ is the (possibly rectangular) diagonal matrix of singular values.   
    Hence, the pseudo-inverse of $\Cmat$ is given by 
        \begin{equation}
            \Cmat\p = \Vmat\T\Lambmat\p\Umat.
        \end{equation}
    Note that the diagonal elements of $\Lambmat\p$ are the reciprocals of the non-zero diagonal values of $\Lambmat$, so we have
        \begin{equation}\label{eqn:Dmat}
            \Dmat = \Lambmat\p\Lambmat = \begin{bmatrix}\eye{r_{\min}}&\bm{0}\\\bm{0}&\bm{0}\end{bmatrix}\in\Rbb^{p_c\times p_c},
        \end{equation}
    where $r_{\min} = \rank(\Cmat)= \min\{n,p_c\}$ is the rank of $\Cmat$. Here, we have used the fact that a random matrix with i.i.d Gaussian entries has full rank with probability (w.p.) 1 \cite{rudelson_non-asymptotic_2010}. In particular,  by \cite[Eqn.~3.2]{rudelson_non-asymptotic_2010}, a square Gaussian  matrix $\Mmat \in \Rbb^{r_{\min} \times r_{\min}}$ is invertible w.p. $1$. Hence, a rectangular Gaussian matrix in $\Rbb^{n \times p_c}$ (which has $\Mmat$ as a sub-matrix) has full rank.   
    
    Hence, $\Cmat\p\Cmat$ can be expressed as follows
        \begin{equation}
            \Cmat\p\Cmat = \Vmat\T\Lambmat\p\Umat\Umat\T\Lambmat\Vmat = \Vmat\T\Lambmat\p\Lambmat\Vmat=\Vmat\T\Dmat\Vmat,
        \end{equation}
    where we have used the fact that for a real orthogonal matrix $\Umat$ we have $\Umat \Umat\T = \eye{}$.
    
    Taking the trace of  $\zvec\T\Ebb[\Cmat\p\Cmat]\zvec$, we obtain
        \begin{align}
            \tr(\zvec\T\Ebb[\Cmat\p\Cmat]\zvec)
          &=  \Ebb[ \tr(  \Dmat \Vmat\zvec\zvec\T\Vmat\T) ] \\
         \label{eqn:DVzzV}
          &=   \tr(  \Dmat  \Ebb [\Vmat\zvec\zvec\T\Vmat\T]) .
        \end{align}
    In \eqref{eqn:DVzzV}, we have moved the expectation inside since $\Dmat$ is given by \eqref{eqn:Dmat} w.p. $1$. 
    Since $\Dmat$ is diagonal and we would like to evaluate a trace type expression, we only need to consider the diagonal elements of $\Ebb[\Vmat\zvec\zvec\T\Vmat\T]$.
    Denoting the $i$\textsuperscript{th} row of $\Vmat$ as $\vvec_i$, the $i$\textsuperscript{th} diagonal element is
        \begin{equation}\label{eqn:squareszv}
               \Ebb[(v_{i,1}z_1+\cdots+v_{i,{p_c}}z_{p_c})^2].
        \end{equation}
    Note that  $\Vmat$ is Haar distributed  since $\Cmat$ is Gaussian distributed \cite{TulinoVerdu_2004}.  
    Hence, by \cite[Lemma~1.1]{hiai_asymptotic_nodate}, the cross terms of the square in \eqref{eqn:squareszv} are zero, and by \cite[Proposition~1.2]{hiai_asymptotic_nodate}, the non-cross terms are $\tfrac{1}{{p_c}}$.
    Hence, we have the $i$\textsuperscript{th} diagonal element of $\Ebb[\Vmat\zvec\zvec\T\Vmat\T]$ as
        \begin{equation}\label{eqn:projection:norm}
            \frac{1}{{p_c}}\norm{\zvec}^2.
        \end{equation}
    We note that \eqref{eqn:projection:norm} does not depend on the order of indices $\zvec$. (This is also a direct consequence of the rotational invariance of Gaussian matrices, i.e.  $\Vmat$, $\Umat$ are Haar distributed.)
    Using \eqref{eqn:projection:norm}, we  express $\zvec\T\Ebb[\Cmat\p\Cmat]\zvec$ as follows as given in Lemma~\ref{lem:wisharts-nice-dream}
    \begin{equation}
        \zvec\T\Ebb[\Cmat\p\Cmat]\zvec = \norm{\zvec}^2\frac{r_{\min}}{{p_c}},
    \end{equation}
    where $r_{\min} = \min\{n,p_c\}$.
\fi

\iftr
    \subsection{Proof of Lemma \ref{lem:A^+A}}\label{sec:pf:lem:A^+A}

    We first focus on $\Gmat= \Amat\T\Bar{\Amat}\T\Bar{\Amat}\Amat$. 
    Using the definition of $\Bar{\Amat}$ in \eqref{eqn:def:Amat}, we express the product $\Bar{\Amat}\T\Bar{\Amat}$ as follows
        \begin{equation}
           \Bar{\Amat}\T\Bar{\Amat} = \sum_{k=1}^K (\Amat_k\Amat_k\T)\p,
        \end{equation}
    where we have used the following identities for the pseudoinverse: $(\Mmat\p)\T = (\Mmat\T)\p$ and $(\Mmat\T)\p\Mmat\p = (\Mmat\Mmat\T)\p$ for any matrix $\Mmat$.
    
    Hence,  we have  
    \begin{align}
    \Gmat =\Amat\T\Bar{\Amat}\T\Bar{\Amat}\Amat = \Amat\T \sum_{k=1}^K (\Amat_k\Amat_k\T)\p  \Amat
    \end{align}
    The matrix $\Gmat$ and hence $\Ebb[\Gmat]$ can be seen as a matrix consisting of $K\times K$ blocks of varying sizes.  The $(k,j)$\textsuperscript{th} block of $\Ebb[\Gmat]$ ($k$\textsuperscript{th} horizontal, $j$\textsuperscript{th} vertical block) is given by  
        \begin{align}\label{eqn:Gblock}
            \Ebb[\Amat_k\T\sum_{i=1}^K(\Amat_i\Amat_i\T)\p\Amat_j].
        \end{align}
    We now consider the cases with $k \neq j$ and $k= j$, separately. 
    For $k\neq j$, \eqref{eqn:Gblock} can be written as 
        \begin{align}
        \begin{split}
            &\Ebb[\Amat_k\T\sum_{i=1}^K(\Amat_i\Amat_i\T)\p\Amat_j]\\
            &= \Ebb[\Amat_k\T(\Amat_k\Amat_k\T)\p\Amat_j + \Amat_k\T(\Amat_j\Amat_j\T)\p\Amat_j \\
            &\quad + \Amat_k\T\sum_{\substack{i=1\\i\neq k\\i\neq j}}^K(\Amat_i\Amat_i\T)\p\Amat_j]
        \end{split}\label{eqn:crazy_sum}\\
               &= \zerobf
        \end{align}
    Here we have used the fact that the rows of $\Amat$ are i.i.d. with  $\sim \Gauss{p}$, hence the matrices $\Amat_j$  and $\Amat_i$ are statistically independent for $i\neq j$.
    Thus, using the fact that $\Ebb[\Amat_l]=0$, $\forall l$,  \eqref{eqn:crazy_sum} is equal to the zero matrix of appropriate dimensions (under $k\neq j$).
 
    For the second case, $k=j$, i.e.,  the $(k,k)$\textsuperscript{th} block is given by
        \begin{equation}
            \Ebb[\Amat_k\T\sum_{i=1}^K(\Amat_i\Amat_i\T)\p\Amat_k].
        \end{equation}
    We now consider the above expression together with the terms including $\zvec$. In particular, partitioning the vector $\zvec$ as $\zvec=[\zvec_1;\,\cdots;\,\zvec_K]$ where $\zvec_k\in\Rbb^{p_k\times 1}$, we obtain
        \begin{align}\label{eqn:double_sum_zAz}
        \begin{split}
             &\zvec\T\Ebb[\Amat\T\Bar{\Amat}\T\Bar{\Amat}\Amat]\zvec \\ &=\sum_{k=1}^K\left(\zvec_k\T\sum_{i=1}^K\Ebb\left[  \Amat_k\T(\Amat_i\Amat_i\T)\p\Amat_k\right]\zvec_k\right).
        \end{split}
        \end{align}
    To evaluate \eqref{eqn:double_sum_zAz}, we will first derive expressions for the terms with $k=i$ and then for the terms with $k\neq i$.
    
    $k=i$: Note that for a matrix $\Mmat$ and its pseudoinverse $\Mmat\p$, we have $\Mmat\T(\Mmat\Mmat\T)\p\Mmat=\Mmat\p\Mmat$. 
    Hence, we obtain 
        \begin{equation}\label{eqn:AT(AAT)+A=A+A}
            \zvec_k\T\Ebb[\Amat_k\T(\Amat_k\Amat_k\T)\p\Amat_k]\zvec_k = \zvec_k\T\Ebb[\Amat_k\p\Amat_k]\zvec_k,
        \end{equation}
    By combining \eqref{eqn:AT(AAT)+A=A+A} with Lemma \ref{lem:wisharts-nice-dream}, we obtain
        \begin{equation}\label{eqn:k=i}
            \zvec_k\T\Ebb[\Amat_k\T(\Amat_k\Amat_k\T)\p\Amat_k]\zvec_k = \norm{\zvec_k}^2 \frac{r_{\min,k}}{p_k},
        \end{equation}
    where $r_{\min,k} = \min\{n,p_k\}$.
    
    $k\neq i$: Given that the rows of $\Amat$ are i.i.d. with  $\sim \Gauss{p}$, the columns are also i.i.d.
    Thus, $\Amat_k$ and $\Amat_i$ are statistically independent for $k\neq i$ and:
        \begin{equation}
            \Ebb[\Amat_k\T(\Amat_i\Amat_i\T)\p\Amat_k] = \Ebb[\Amat_k\T\Ebb[(\Amat_i\Amat_i\T)\p]\Amat_k].
        \end{equation}
    Following the notation of \cite{cook_mean_2011}, $\Amat_i\Amat_i\T$ follows the $n$-variate Wishart distribution with $p_k$ degrees of freedom: $W_n(\eye{p_k},p_k)$.
    The pseudoinverse $(\Amat_i\Amat_i\T)\p$ follows the inverse Wishart distribution if $p_k > n + 1 $, and the generalized inverse Wishart distribution if $p_k < n - 1$ \cite{cook_mean_2011}.
    From \cite{cook_mean_2011} we have the following expression for $\Ebb[(\Amat_i\Amat_i\T)\p]$:
        \begin{equation}\label{eqn:EAATp}
            \Ebb[(\Amat_i\Amat_i\T)\p]= \gamma_k' \eye{n},
        \end{equation}
    where
        \begin{numcases}{\gamma_k'=}
            \tfrac{1}{p_k-n-1}         &\hspace{-15pt} for $p_k > n + 1$, \\
            \tfrac{p_k}{n(n - p_k - 1)}  &\hspace{-15pt} for $p_k < n - 1$, \\
            +\infty &\hspace{-15pt}  for $p_k \in \{n-1, n, n+1\}.$
        \end{numcases}
    Note that the more restrictive conditions on $p_k,n$ in \cite{cook_mean_2011} is due to the fact Prop.2.1\, and Thm~2.1 of \cite{cook_mean_2011} also present the second order moments for which more restrictive conditions are needed.  
    Hence, we have
        \begin{equation}
            \Ebb[\Amat_k\T(\Amat_i\Amat_i\T)\p\Amat_k] = \gamma_k'\Ebb[\Amat_k\T\Amat_k].
        \end{equation}
    The columns of $\Amat_k$ are i.i.d. standard Gaussian of dimension $n\times 1$, so we have
        \begin{equation}\label{eqn:kneqi}
            \gamma_k'\Ebb[\Amat_k\T\Amat_k] = \gamma_k' n \eye{p_k} = \gamma_k\eye{p_k},
        \end{equation}
    where $\gamma_k$'s are as defined in \eqref{eqn:gamma_main}. 
    
    Combining \eqref{eqn:k=i} and \eqref{eqn:kneqi} with \eqref{eqn:double_sum_zAz} we obtain the desired equality
        \begin{equation}
            \zvec\T\Ebb[\Amat\T\Bar{\Amat}\T\Bar{\Amat}\Amat]\zvec = \sum_{k=1}^K \norm{\zvec_k}^2 \left( \tfrac{r_{\min,k}}{p_k} + \sum_{\substack{i=1\\i\neq k}}^K \gamma_i \right).
        \end{equation}
    This concludes the proof of Lemma~\ref{lem:A^+A}.
\fi

\bibliographystyle{IEEEtran}
\bibliography{ref}

\begin{thebibliography}{10}
\providecommand{\url}[1]{#1}
\csname url@samestyle\endcsname
\providecommand{\newblock}{\relax}
\providecommand{\bibinfo}[2]{#2}
\providecommand{\BIBentrySTDinterwordspacing}{\spaceskip=0pt\relax}
\providecommand{\BIBentryALTinterwordstretchfactor}{4}
\providecommand{\BIBentryALTinterwordspacing}{\spaceskip=\fontdimen2\font plus
\BIBentryALTinterwordstretchfactor\fontdimen3\font minus
  \fontdimen4\font\relax}
\providecommand{\BIBforeignlanguage}[2]{{%
\expandafter\ifx\csname l@#1\endcsname\relax
\typeout{** WARNING: IEEEtran.bst: No hyphenation pattern has been}%
\typeout{** loaded for the language `#1'. Using the pattern for}%
\typeout{** the default language instead.}%
\else
\language=\csname l@#1\endcsname
\fi
#2}}
\providecommand{\BIBdecl}{\relax}
\BIBdecl

\bibitem{verbraeken_survey_2019}
J.~{Verbraeken}, M.~{Wolting}, J.~{Katzy}, J.~{Kloppenburg} \emph{et~al.}, ``{A
  Survey on Distributed Machine Learning},'' \emph{arXiv:1912.09789}, Dec 2019.

\bibitem{niknam_federated_2019}
S.~{Niknam}, H.~S. {Dhillon}, and J.~H. {Reed}, ``{Federated Learning for
  Wireless Communications: Motivation, Opportunities and Challenges},''
  \emph{arXiv:1908.06847}, Jul 2019.

\bibitem{wang2020convergence}
X.~Wang, Y.~Han, V.~C. Leung, D.~Niyato \emph{et~al.}, ``Convergence of edge
  computing and deep learning: A comprehensive survey,'' \emph{IEEE
  Communications Surveys \& Tutorials}, 2020.

\bibitem{zhang_understanding_2017}
C.~{Zhang}, S.~{Bengio}, M.~{Hardt}, B.~{Recht} \emph{et~al.}, ``{Understanding
  deep learning requires rethinking generalization},'' \emph{arXiv:1611.03530},
  Nov 2016.

\bibitem{predd_distributed_2006}
J.~B. {Predd}, S.~B. {Kulkarni}, and H.~V. {Poor}, ``Distributed learning in
  wireless sensor networks,'' \emph{IEEE Signal Processing Magazine}, vol.~23,
  no.~4, pp. 56--69, July 2006.

\bibitem{magnusson_communication_2018}
S.~{Magnússon}, C.~{Enyioha}, N.~{Li}, C.~{Fischione} \emph{et~al.},
  ``Communication complexity of dual decomposition methods for distributed
  resource allocation optimization,'' \emph{IEEE Journal of Selected Topics in
  Signal Processing}, vol.~12, no.~4, pp. 717--732, Aug 2018.

\bibitem{tsitsiklis_distributed_1984}
J.~{Tsitsiklis}, D.~{Bertsekas}, and M.~{Athans}, ``Distributed asynchronous
  deterministic and stochastic gradient optimization algorithms,'' \emph{IEEE
  Trans. on Automatic Control}, vol.~31, no.~9, pp. 803--812, Sep. 1986.

\bibitem{smith_cocoa_nodate}
V.~Smith, S.~Forte, C.~Ma, M.~Tak{\'a}{\v{c}} \emph{et~al.}, ``Cocoa: A general
  framework for communication-efficient distributed optimization,''
  \emph{Journal of Machine Learning Research}, vol.~18, pp. 1--49, 2018.

\bibitem{belkin_reconciling_2019}
M.~Belkin, D.~Hsu, S.~Ma, and S.~Mandal, ``Reconciling modern machine-learning
  practice and the classical bias{\textendash}variance trade-off,'' \emph{Proc.
  of the National Academy of Sciences}, vol. 116, no.~32, pp. 15\,849--15\,854,
  2019.

\bibitem{belkin_two_2019}
M.~{Belkin}, D.~{Hsu}, and J.~{Xu}, ``{Two models of double descent for weak
  features},'' \emph{arXiv:1903.07571}, Mar 2019.

\bibitem{he_cola_2019}
L.~He, A.~Bian, and M.~Jaggi, ``Cola: Decentralized linear learning,''
  \emph{Advances in Neural Information Process. Systems}, pp. 4536--4546, 2018.

\bibitem{breiman_how_nodate}
L.~Breiman and D.~Freedman, ``How many variables should be entered in a
  regression equation?'' \emph{Journal of the American Statistical
  Association}, vol.~78, no. 381, pp. 131--136, 1983.

\bibitem{b_HaykinAdaptive}
S.~Haykin, \emph{Adaptive Filter Theory}.\hskip 1em plus 0.5em minus
  0.4em\relax Prentice-Hall, Inc., 1996.

\bibitem{rudelson_non-asymptotic_2010}
M.~Rudelson and R.~Vershynin, ``\BIBforeignlanguage{en}{Non-asymptotic theory
  of random matrices: extreme singular values},''
  \emph{\BIBforeignlanguage{en}{arXiv:1003.2990}}, Apr. 2010.

\bibitem{TulinoVerdu_2004}
A.~M. Tulino and S.~Verd\'u, ``Random matrix theory and wireless
  communications,'' \emph{Foundations and Trends in Communications and
  Information Theory}, 2004.

\bibitem{hiai_asymptotic_nodate}
F.~Hiai and D.~Petz, ``\BIBforeignlanguage{en}{Asymptotic {Freeness} {Almost}
  {Everywhere} for {Random} {Matrices}},'' \emph{\BIBforeignlanguage{en}{Acta
  Sci. Math. (Szeged)}}, vol.~66, pp. 801--826, 2000.

\bibitem{cook_mean_2011}
R.~D. Cook and L.~Forzani, ``On the mean and variance of the generalized
  inverse of a singular {Wishart} matrix,'' \emph{Electron. J. Statist.},
  vol.~5, pp. 146--158, 2011.

\end{thebibliography}

\end{document}